\documentclass[letterpaper]{article} 
\usepackage{aaai25}  
\usepackage{times}  
\usepackage{helvet}  
\usepackage{courier}  
\usepackage[hyphens]{url}  
\usepackage{graphicx} 
\usepackage{natbib}  
\usepackage{caption} 
\usepackage{algorithm}
\usepackage{algorithmic}

\usepackage{amsmath}
\usepackage{dsfont}
\usepackage{multirow}
\usepackage{bbding}

\usepackage{newfloat}
\usepackage{listings}
\DeclareCaptionStyle{ruled}{labelfont=normalfont,labelsep=colon,strut=off} 
\floatstyle{ruled}
\newfloat{listing}{tb}{lst}{}
\floatname{listing}{Listing}
\title{Multi-Attribute Multi-Grained Adaptation of Pre-Trained Language Models for Text Understanding from Bayesian Perspective}
\author{
    You Zhang\textsuperscript{\rm 1},
    Jin Wang\textsuperscript{\rm 1}\equalcorreauthor,
    Liang-Chih Yu\textsuperscript{\rm 2}\equalcorreauthor,
    Dan Xu\textsuperscript{\rm 1},
    Xuejie Zhang\textsuperscript{\rm 1}\\
}
\affiliations{
    \textsuperscript{\rm 1}School of Information Science and Engineering, Yunnan University, Yunnan, P.R.China\\
    \textsuperscript{\rm 2}Department of Information Management, Yuan Ze University, Taiwan\\
    \{yzhang0202, wangjin\}@ynu.edu.cn, lcyu@saturn.yzu.edu.tw

}

\usepackage{bibentry}

\begin{document}

\maketitle

\begin{abstract}
Current neural networks often employ multi-domain-learning or attribute-injecting mechanisms to incorporate non-independent and identically distributed (non-IID) information for text understanding tasks by capturing individual characteristics and the relationships among samples. However, the extent of the impact of non-IID information and how these methods affect pre-trained language models (PLMs) remains unclear. This study revisits the assumption that non-IID information enhances PLMs to achieve performance improvements from a Bayesian perspective, which unearths and integrates non-IID and IID features. Furthermore, we proposed a multi-attribute multi-grained framework for PLM adaptations (M2A), which combines multi-attribute and multi-grained views to mitigate uncertainty in a lightweight manner. We evaluate M2A through prevalent text-understanding datasets and demonstrate its superior performance, mainly when data are implicitly non-IID, and PLMs scale larger.
\end{abstract}

\begin{links}
    \link{Code}{https://github.com/yoyo-yun/M2A.}
\end{links}

\begin{figure*}[t!]
\centering
\includegraphics[width=4.2in]{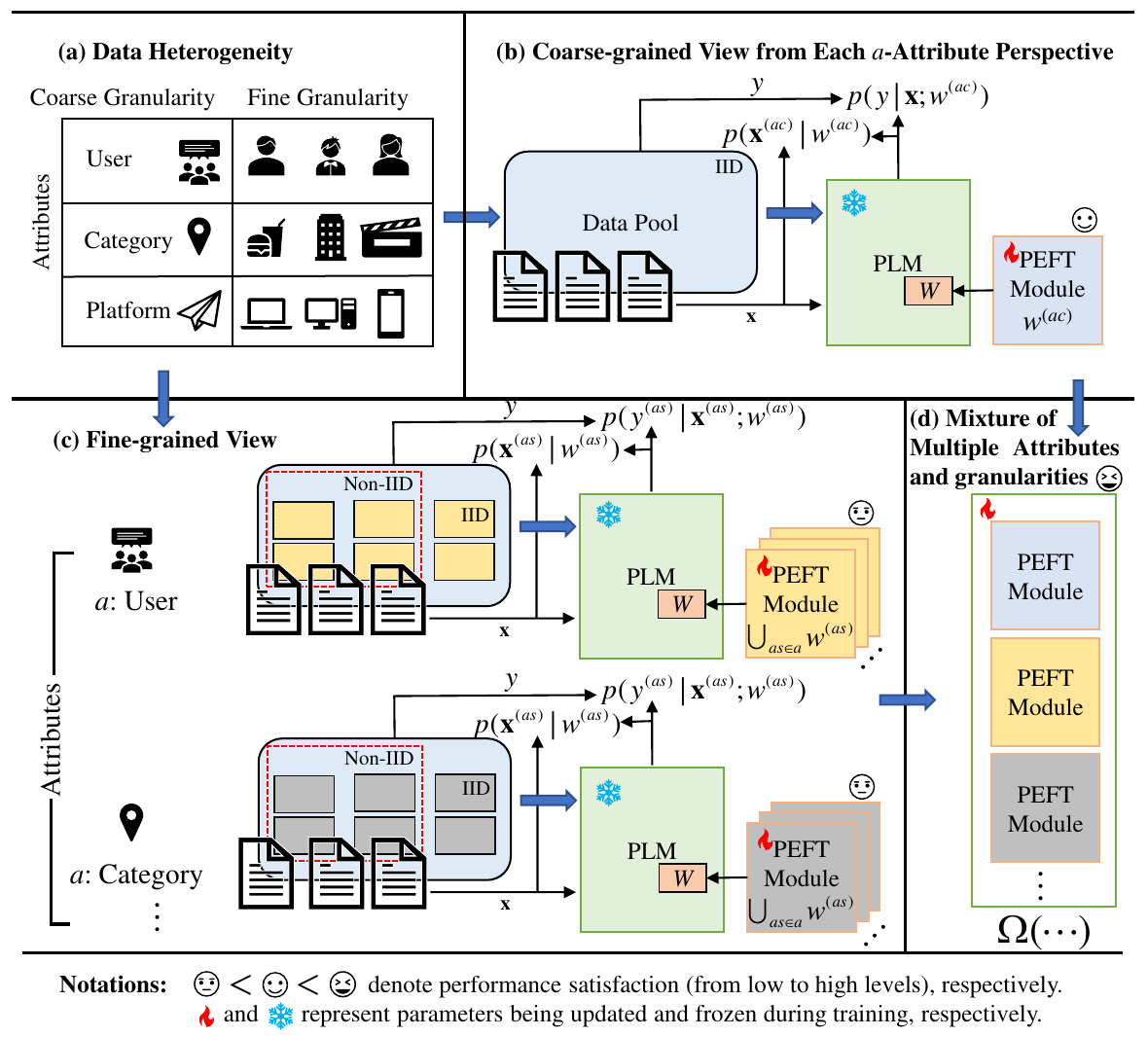}
\caption{A conception of the proposed method.} \label{fig:1}
\end{figure*}

\section{Introduction}
Modern neural networks, especially pre-trained language models (PLMs), for specific tasks require large amounts of data and computational resources~\cite{Kong2021,Han2021}; therefore, collecting as much data as possible from diverse sources to achieve satisfactory task performance is desirable~\cite{Yuan2022,Katsarou2023,Liu2018,Yang2015}. For instance, sentiment analysis data can be gathered from different users, categories, and platforms, where corresponding golden labels are annotated by different users and intended to build sufficient samples leveraged for robust sentiment model training~\cite{,Chen2021,Zhang2023}. These networks unanimously assume that all samples in a data pool (for a specific downstream task) are independent and identically distributed (IID) in a coarse-grained view~\cite{Pang2006a,Zhang2018}. However, data distributions from different resources are not typically IID, which needs fine-grained view to deal with the data heterogeneity problem. Without the fine-grained view, the coarse-grained IID assumption may impede the optimal model convergences (uncertainty), degrading ideal performance when data are non-IID, either explicitly or implicitly, even though carrying out sufficient data~\cite{Zhang2023,Yao2024}.

One intuitive solution to address the issue is to abstract the heterogeneities in data-sufficient while resource-diverse tasks and guide models adapted to diverse individual scenarios. This brings two significant challenges in recent adaptations of PLMs: 1) \textbf{representing the heterogeneities among data samples} and 2) \textbf{making PLMs capable of recognizing them effectively and efficiently in the training and inference phase}. Recently, many efforts have been committed to introducing multi-domain-learning~\cite{Yuan2022,Katsarou2023}, and attribute-injecting mechanisms~\cite{Zhang2021,Amplayo2019} for investigating data heterogeneities. Multi-domain learning aims to learn domain-shared features and combine them with separate domain-specific features for multi-domain text classifications. Attribute injecting treats concrete domain information as attribute knowledge and injects it into coarse neural networks for fine-grained performance. However, these high-performance methods do not systematically develop the model design and optimization approach, constraining the ability to represent robust non-IID and IID features as well as their internal relatedness. Moreover, traditional full-model fine-tuning (FFT) and sophisticated structure modification render limited scalability for larger PLMs~\cite{Min2021}.

We argue that most text-understanding data contain multiple attributes, i.e., user, category, and platform. From a data distribution perspective, each attribute covers coarse-grained and fine-grained views, as shown in Figure~\ref{fig:1}(a). The coarse-grained view with one domain, which treats all samples as IID, ignores data heterogeneities. This approach is advantageous for accommodating large number of samples, as shown in Figure~\ref{fig:1}(b). The fine-grained view, e.g., the category, considers all samples divided into numerous fine-grained domains. Within each domain, samples are IID, whereas samples across different domains are non-IID. This perspective allows models to manage diverse samples by employing individual modules for each domain, as shown in Figure~\ref{fig:1}(c).

To integrate IID and non-IID information for complementary benefits, we rethink the current multi-domain learning and attribute injecting mechanism and propose a multi-attribute, multi-grained adaptation (M2A) framework for mitigating the uncertainty of PLMs’ adaptation, as shown in Figure~\ref{fig:1}(d). Theoretically, we adopt a Bayesian inference to analyze the relatedness between coarse-grained and fine-grained views and a Bayesian neural network (BNN) to represent robust data heterogeneities~\cite{Magris2023,Jospin2022} (see §Preliminaries). Through a combination of multiple attributes and granularities from the Bayesian inference perspective, we found that the proposed M2A framework could facilitate PLMs to be effectively adapted to non-IID tasks. Moreover, a parameter-efficient fine-tuning (PEFT) module~\cite{Houlsby2019}  and a joint learning strategy are proposed to facilitate the scalability of M2A for diverse PLMs~\cite{Zhang2023}. In experiments, we utilized various PLMs to evaluate M2A on multi-domain and personalized sentiment analysis tasks, which contain sufficient data for task adaptation while involving data heterogeneities.

Our key contributions in this paper are threefold.
\begin{itemize}
\item We rethink the current multi-domain-learning and attribute-injecting mechanism from a Bayesian perspective and provide a robust M2A framework for text understanding.
\item Based on PEFT and joint learning methods, M2A makes PLMs effectively and efficiently adapted for diverse scales and broader scenarios.
\item Extensive experiments were conducted on several text-understanding tasks and provided empirical analysis.
\end{itemize}

\begin{figure*}[t!]
\centering
\includegraphics[width=5.8in]{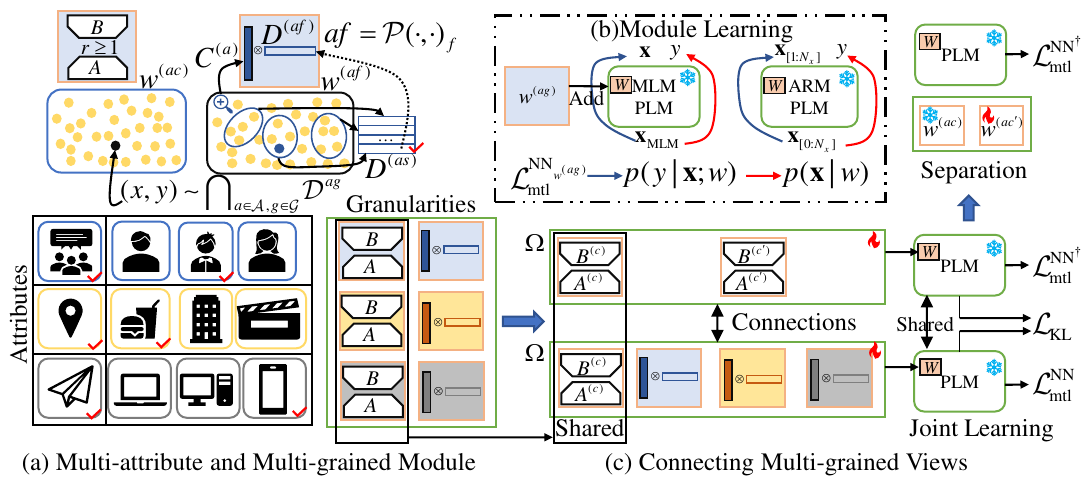}
\caption{Overview of the M2A Framework.} \label{fig:2}
\end{figure*}

\section{Preliminaries} \label{sec:2}
\subsection{Background: Text Understanding} \label{sec:2.1}
Regarding a text understanding task with a training dataset ${{\cal D}} = \{ ({{\bf{x}}_n},{y_n})\} _n^N$ with $N$ IID and label-annotated samples, a neural network ${\rm{N}}{{\rm{N}}_w}( \cdot )$ parameterized with weights $w$ is optimized to estimate the likelihood $p({{{\cal D}}_y}|{{{\cal D}}_x};w)$ that links the inputs and the outputs via:
\begin{equation}
p({{{\cal D}}_y}|{{{\cal D}}_x};w) = \mathop {\arg \max }\limits_w \prod\limits_{n = 1}^N {{\rm{Loss}}({y_n},{{\hat y}_n})} \label{eq:1}
\end{equation}
where ${{{\cal D}}_x} = \{ {{\bf{x}}_n}\} _{n = 1}^N$ and ${{{\cal D}}_y} = \{ {y_n}\} _{n = 1}^N$ represents the input data and the output labels; ${\hat y_n} = {\rm{N}}{{\rm{N}}_w}({{\bf{x}}_n})$ denotes the network output for input ${{\bf{x}}_n}$; ${\rm{Loss}}( \cdot )$ the loss function for model optimizations.

Regarding non-IID heterogeneities in a specific attribute $a \in {{\cal A}}$, the dataset ${{\cal D}} = { \cup _{as \in a}}{{{\cal D}}^{(as)}}$ can be divided into $|a|_f$ fine-grained domains (sub-datasets) where each domain contains IID samples, denoted as ${{{\cal D}}^{(as)}} = \{ ({\bf{x}}_n^{(as)},y_n^{(as)})\} _n^{|as|}$. For each domain $as$, it typically requires a neural network ${\rm{N}}{{\rm{N}}_{{w^{(as)}}}}( \cdot )$ to estimate the domain-specific sub-datasets. Notably, only one IID domain $ac$ is covered in coarse-grained view, where ${{\cal D}} = {{{\cal D}}^{(ac)}}$, and the corresponding model is ${\rm{N}}{{\rm{N}}_{{w^{(ac)}}}}( \cdot )$.

\subsection{A Bayesian Perspective} \label{sec:2.2}
\textbf{Bayesian Inference.} We propose a Bayesian perspective to investigate diversity and relatedness among multiple attributes and granularities, respectively. In Bayesian learning~\cite{Freitas2000}, $p(w|{{\cal D}})$ is the posterior distribution of the model being learned. Therefore, the predictive probabilities can be defined as:
\begin{equation}
p(y|{\bf{x}};{{\cal D}}) = \int {p(y|{\bf{x}};w)p(w|{{\cal D}}){d_w}}  \label{eq:2}
\end{equation}
which can be approximated by Monte Carlo method~\cite{kroese2014monte} and aggregated from $|S|$ models:
\begin{equation}
p(y|{\bf{x}};{{\cal D}}) \approx {1 \over {|S|}}\sum\limits_{s \in S} {p(y|{\bf{x}};{w^{(s)}})} \label{eq:3}
\end{equation}
where ${w^{(s)}} \sim p({w^{(s)}}|{{\cal D}})$.

\noindent \textbf{Bayesian Neural Network.}
It targets the estimation of posterior distribution $p(w|{{\cal D}})$ based on a Bayesian theorem via~\cite{Jospin2022}:
\begin{equation}
\begin{aligned}
p(w|{{{\cal D}}_x},{{{\cal D}}_y}) &= {{p({{{\cal D}}_y}|{{{\cal D}}_x};w) \cdot p({{{\cal D}}_x}|w) \cdot p(w)} \over {p({{{\cal D}}_x},{{{\cal D}}_y})}} \cr
&\propto p({{{\cal D}}_y}|{{{\cal D}}_x};w) \cdot p({{{\cal D}}_x}|w) \label{eq:4}
\end{aligned}
\end{equation}
where $p({{{\cal D}}_x}|w)$ is the likelihood of the model that predicts input data; $p(w)$ is the prior distribution of the model; $p({{\cal D}}) = p({{{\cal D}}_x},{{{\cal D}}_y})$ is the evidence.

To sum up, the question of our proposed method is: how to (1) integrate multiple attributes and granularities for robust text representations, (2) estimate $p(w|{{\cal D}})$ in each domain for capturing the non-IID and IID features, and (3) connect non-IID and IID information for modeling the internal relatedness.

\section{M2A Framework} \label{sec:3}
Figure~\ref{fig:2} illustrates the conceptual diagram of the proposed M2A method, which consists of three steps from a Bayesian perspective. The first step views each data from different attribute and granularity perspectives, utilizing M2A modules to integrate non-IID and IID features; the second step optimizes M2A modules through multitask learning; and the third step connects multi-grained views and introduces a joint learning strategy. Additionally, our framework can be decomposed to improve parameter efficiency.

\subsection{Multi-attribute and Multi-grained Module in Bayesian Inference} \label{sec:3.1}
Based on Bayesian inference in Eq. (\ref{eq:3}), we introduce a multi-attribute and multi-grained module to represent and integrate non-IID and IID information for robust hidden representation.

Each data sample $({\bf{x}},y)$ can be viewed from different attributes $a \in {{\cal A}}$ and multi-grained views $g \in {{\cal G}} = \{ c,f\} $ and denoted by $({\bf{x}},y) \sim \bigcap\nolimits_{a \in {{\cal A}},g \in {{\cal G}}} {{{{\cal D}}^{(ag)}}} $, where $af$ represents $({\bf{x}},y)$ in the fine-grained domain $f$ from $a$-attribute perspective\footnote{For simplification, we argue each data sample $({\bf{x}},y)$ belongs to only one fine-grained domain $f \in a$ in each attribute view $a$.} via the formulation of $af = {{\cal P}}{(({\bf{x}},y),a)_f}$. In contrast to Monte Carlo sampling, we sample ${w^{(s)}}$ from diverse data distributions, i.e., ${{{\cal D}}^{(ag)}}$ with $a \in {{\cal A}}$ and $g \in {{\cal G}}$; therefore, the predictive probabilistic of $({\bf{x}},y)$ can be formulated as:
\begin{equation}
p(y|{\bf{x}};{{\cal D}}) \approx {1 \over {\left| {{\cal A}} \right|\left| {{\cal G}} \right|}}\sum\limits_{a \in {{\cal A}},g \in {{\cal G}}} {p(y|{\bf{x}};{w^{(ag)}})} \label{eq:5}
\end{equation}
where ${w^{(ag)}} \sim p({w^{(ag)}}|{{{\cal D}}^{(ag)}})p({{{\cal D}}^{(ag)}}|{{\cal D}})$. Since these $w^{(ag)}$s have complementary and individual semantic information, $w^{(ag)}$s can be fused to a unified model in current models~\cite{zhang2024personalized}:
\begin{equation}
p(y|{\bf{x}};{{\cal D}}) \approx p(y|{\bf{x}};\Omega (\bigcup\nolimits_{ag} {{w^{(ag)}}} )) \label{eq:6}
\end{equation}
where $\Omega ( \cdots )$ represents module integrating operations.

Inspired by PEFT methods that introduce a lightweight module to enable PLMs to be adapted to downstream tasks, we utilize the LoRA family to implement our M2A framework for multi-objective adaptations efficiently, facilitating further research when PLMs scale large~\cite{Hu2021}. Each module ${w^{(ag)}}$ is formulated as ${A^{(ag)}}{B^{(ag)}}$ with low-rank instinct features of ${r^{(ag)}} \ll \min ({d_{in}},{d_{out}})$, which are then added into PLM matrix parameters $W \in {\mathds{R}^{{d_{in}} \times {d_{out}}}}$ for model calibration:
\begin{equation}
W = W + {w^{(ag)}} \label{eq:7}
\end{equation}

Due to the cumulative and plug-and-play characteristics of LoRA~\cite{wu2023mole}, we instantiate modules integrating operations as $\Omega (\bigcup\nolimits_{a \in {{\cal A}},g \in {{\cal G}}} {{w^{(ag)}}} ) = {1 \over {\left| {{\cal A}} \right|\left| {{\cal G}} \right|}}\sum\nolimits_{ag} {{w^{(ag)}}} $. After that, as shown in Figure~\ref{fig:2}, our M2A module is simplified and denoted as:
\begin{equation}
W = W + {1 \over {\left| {{\cal A}} \right|\left| {{\cal G}} \right|}}\sum\limits_{a \in {{\cal A}},g \in {{\cal G}}} {{w^{(ag)}}} \label{eq:8}
\end{equation}

\subsection{Module Learning in BNN} \label{sec:3.2}
The essential factor in optimizing the module $w$ locates how to maximize $p(w|{{{\cal D}}_x},{{{\cal D}}_y})$  based on PLM backbones. 

Although PLMs have provided general knowledge for text generation or cloze tasks, they could not effectively distinguish the data heterogeneities. For example, due to different pragmatic writing styles, PLMs make it hard to determine word prediction for different writers with the same given texts. Based on Eq. (\ref{eq:4}) with the Bayesian diagram, we introduce a multitask learning to estimate $p(w|{{\cal D}})$ via maximizing $p({{{\cal D}}_y}|{{{\cal D}}_x};w)$ and $p({{{\cal D}}_x}|w)$, simultaneously:
\begin{equation}
{{\cal L}}_{{\rm{mtl}}}^{{\rm{N}}{{\rm{N}}_w}} = {\rm{Loss}}({y_n},{\hat y_n}) + \alpha {\rm{Loss}}({{\bf{x}}_n},{{\bf{\hat x}}_n}) \label{eq:9}
\end{equation}
where $\alpha $ is the balance factor; ${{\bf{\hat x}}_n}$ denote reconstructed text tokens under $p({{{\cal D}}_x}|w)$ estimation and  ${\rm{N}}{{\rm{N}}_w}( \cdot )$. Empirically, from the writer-specific perspective (i.e., $a$ denotes writer attribute), $p({{{\cal D}}_y}|{{{\cal D}}_x};{w^{(as)}})$ aims to facilitate PLMs to learn \textit{how a specific writer (i.e., s) would influence the annotations of the texts}. Meanwhile, $p({{{\cal D}}_x}|{w^{(as)}})$ is introduced to optimize PLMs to learn \textit{What pragmatic contents a specific writer (i.e., s) would generate}.

To implement $p({{{\cal D}}_x}|w)$ based on masked language modeling (MLM)~\cite{Devlin2019} and autoregressive modeling (ARM)~\cite{radfordlanguage}, respectively, we utilized different strategies. For MLM, we randomly mask 15\% of input tokens as \texttt{[MASK]} and get unified models to predict original tokens of texts and annotated labels based on \texttt{[CLS]} or \texttt{<s>} tokens. For ARM-based PLMs, ARM tasks are used to predict both shift-right text labels and downstream task labels.

\subsection{Connecting Multi-grained Views} \label{sec:3.3}
Acknowledging that identifying exact non-IID information locations could facilitate neural network models to mitigate predictive uncertainty, it is difficult to collect the locations due to extensive labor in retrieving sensitive information and bypassing privacy concerns~\cite{Zhang2023}. To further investigate internal relatedness among coarse-grained and fine-grained views, Bayesian inference in Eq. (\ref{eq:3}) guides the M2A framework with a joint learning strategy for improving the model performance and broadening practical scenarios. A more detailed analysis can be found in the technical appendices. To generalize fine-grained domains, we can get a multi-attribute coarse-grained M2A module (dubbed as M2A†), denoted by $W + \Omega (\bigcup\nolimits_{a \in {{\cal A}},g \in \{ c,c'\} } {{w^{(ag)}}} )$, where $w^{(ac')}$ is a special coarse-view module used to align diverse fine-grained modules.

To align with M2A and M2A†, we resort to a joint learning strategy with knowledge distillation for final optimizations~\cite{Aguilar2020}, formally:
\begin{equation}
{{\cal L}} = \prod\limits_{({\bf{x}},y) \in {{\cal D}}} {{{\cal L}}_{{\rm{mtl}}}^{{\rm{NN}}} + {{\cal L}}_{{\rm{mtl}}}^{{\rm{N}}{{\rm{N}}^\dag }} + {\rm{KL}}({\rm{NN}}({\bf{x}}),{\rm{N}}{{\rm{N}}^\dag }({\bf{x}}))} \label{eq:10}
\end{equation}
where ${\rm{NN}} = {\rm{N}}{{\rm{N}}_{W + \Omega (\bigcup\nolimits_{a \in {{\cal A}},g \in \{ c,f\} } {{w^{(ac)}}} )}}$ and ${\rm{N}}{{\rm{N}}^\dag } = {\rm{N}}{{\rm{N}}_{W + \Omega (\bigcup\nolimits_{a \in {{\cal A}},g \in \{ c,c'\} } {{w^{(ag)}}} )}}$. During the training phase, only parameters of $\bigcup\nolimits_{a \in {{\cal A}},g \in \{ c,c',f\} } {{w^{(ag)}}} $ are simultaneously updated. Moreover, we have found that NN converged faster than NN† (see §Experiments); therefore, we introduce a module separation strategy. That means when the learning procedure of NN is optimally stopped during the training phase, NN† will continue learning with only lightweight $\bigcup\nolimits_{a \in {{\cal A}},g \in \{ c'\} } {{w^{(ag)}}} $ to being updated until its optima convergence.

\subsection{Model Decomposition} \label{sec:3.4}
Due to uncertainties in model capacity, the number of domains employed, and module redundancy, we further improve the M2A framework in parameter efficiency via model decomposition (see Figure~\ref{fig:2}). 

\noindent \textbf{Fine-grained domain parameters reduction.}
Regarding model capacity, the fine-grained module (${w^{(af)}}$) covers much fewer data samples than the coarse-grained module (${w^{(ac)}}$ or ${w^{(ac')}}$); thus, we introduce a KronA module~\cite{Edalati2022}, which only contains one-rank weight parameters and provides feasible decompositions, to replace the original LoRA for fine-grained modules, i.e., ${w^{(af)}} = {C^{(af)}} \otimes {D^{(af)}}$ where $ \otimes $ is Kronecker product, ${C^{(af)}} \in {\mathds{R}^{({d_{in}}/r') \times r'}}$, and ${D^{(af)}} \in {\mathds{R}^{r' \times ({d_{out}}/r')}}$ with $1 \le r' \le \min ({d_{in}},{d_{out}})$.

\noindent \textbf{Sharing attribute information among fine-grained domains.}
From a fine-grained perspective, global data can be divided into many domains, which linearly scale model parameters and memory budgets. Here, we share one ${C^{(a)}}$ with all $\bigcup\nolimits_{as \in a} {{C^{(as)}}} $.

\noindent \textbf{Sharing coarse-grained attribute information across attributes.}
Across multiple attributes, the coarse-grained views cover the same global dataset, i.e., ${\forall _{a \in {{\cal A}}}}{{{\cal D}}^{(ac)}} = {{\cal D}}$. To reduce redundancy information, we share one ${w^{(c)}}$ with all $\bigcup\nolimits_{a \in {{\cal A}}} {{w^{(ac)}}} $.

\noindent \textbf{Efficiency analysis.}
Without mode decomposition, the external complexity (ignoring $W$ in PLMs) is $\sum\nolimits_a {({{\left| a \right|}_f}}  + 2) \cdot r \cdot ({d_{in}} + {d_{out}})$, which is larger than that of our decomposition version with $(\sum\nolimits_a {({d_{in}} + {\left| a \right|}_f} \cdot {d_{out}})) + 2 \cdot r \cdot ({d_{in}} + {d_{out}})$.

\begin{table*}[!t]
\small
\centering
\begin{tabular}{|c|c|c|c|c|c|c|c|c|}
\hline
\textbf{Domains} & BERT  & DAEA           & BERTMasker & KCL-KB & B-MTL & B-M2A‡         & G-M2A‡ & R-M2A‡         \\ \hline
Books            & 87.00 & 89.00          & 93.00      & 93.08  & 94.75 & 94.75          & 93.50  & \textbf{97.25} \\ \hline
Electronics      & 88.30 & 91.80          & 93.25      & 94.92  & 94.00 & 95.50          & 93.75  & \textbf{96.00} \\ \hline
DVD              & 85.60 & 88.30          & 89.25      & 89.92  & 90.75 & 93.25          & 91.50  & \textbf{93.50} \\ \hline
Kitchen          & 91.00 & 90.30          & 90.75      & 92.50  & 92.00 & 93.25          & 95.25  & \textbf{96.00} \\ \hline
Apparel          & 90.00 & 89.00          & 92.25      & 92.67  & 91.25 & 92.50          & 93.25  & \textbf{94.50} \\ \hline
Camera           & 90.00 & 92.00          & 92.75      & 93.67  & 94.75 & 95.00          & 93.00  & \textbf{95.25} \\ \hline
Health           & 88.30 & 89.80          & 95.25      & 95.67  & 94.25 & 96.50          & 95.50  & \textbf{97.50} \\ \hline
Music            & 86.80 & 88.00          & 89.50      & 90.42  & 90.75 & 91.75          & 92.00  & \textbf{93.75} \\ \hline
Toys             & 91.30 & 91.80          & 93.75      & 93.33  & 92.75 & 93.50          & 92.25  & \textbf{94.25} \\ \hline
Video            & 88.00 & 92.30          & 91.25      & 91.67  & 92.00 & \textbf{94.50} & 91.75  & 94.00          \\ \hline
Baby             & 91.50 & 92.30          & 92.75      & 94.58  & 95.75 & \textbf{96.25} & 94.00  & \textbf{96.25} \\ \hline
Magazines        & 92.80 & \textbf{96.50} & 94.50      & 94.17  & 94.25 & 94.75          & 94.00  & \textbf{96.50} \\ \hline
Software         & 89.30 & 92.80          & 93.00      & 94.33  & 95.75 & \textbf{96.50}  & 96.25  & \textbf{96.50} \\ \hline
Sports           & 90.80 & 90.80          & 92.50      & 94.42  & 94.25 & \textbf{95.00} & 94.50  & 94.75          \\ \hline
IMDB             & 85.80 & 90.80          & 86.00      & 90.83  & 93.00 & 93.25          & 92.25  & \textbf{94.50} \\ \hline
MR               & 74.80 & 77.00          & 83.75      & 85.58  & 84.25 & \textbf{85.75} & 84.00  & 85.50          \\ \hline
Avg.             & 88.16 & 90.16          & 91.47      & 92.62  & 92.78 & 93.86          & 92.92  & \textbf{94.70} \\ \hline
\end{tabular}
\caption{Comparative test Acc (\%) results for multi-domain sentiment analysis (the category attribute only).}
\label{tab:1}
\end{table*}

\begin{table*}[!t]
\centering
\small
\begin{tabular}{|l|c|c|c|c|c|c|c|c|c|}
\hline
\multicolumn{1}{|c|}{\multirow{2}{*}{\textbf{Models}}} & \multicolumn{3}{c|}{IMDB}                                  & \multicolumn{3}{c|}{Yelp-2013}                  & \multicolumn{3}{c|}{Yelp-2014}                  \\ \cline{2-10}
\multicolumn{1}{|c|}{}                                 & Acc                      & RMSE           & F1            & Acc           & RMSE           & F1            & Acc           & RMSE           & F1            \\\hline
\multicolumn{10}{|c|}{non-IID free or Coarse-grained   view (IID) only}                                                                                                                                              \\\hline
BERT                                                 & 52.2                     & 1.163          & 49.3          & 67.7          & 0.628          & \underline{65.5}    & 67.7          & 0.615          & 65.6          \\\hline
GPT2                                                 & 51.5                     & 1.222          & 47.5          & 67.6          & 0.622          & 64.5          & 68.0          & 0.614          & 65.5          \\\hline
RoBERTa                                              & \underline{53.0}               & \underline{1.147}    & \underline{49.4}    & \underline{69.2}    & \underline{0.590}    & 65.1          & \underline{69.0}      & \underline{0.601}    & \underline{66.5}    \\\hline
R-M2A†                                               & \textbf{54.2}            & \textbf{1.100}   & \textbf{50.3} & \textbf{70.7} & \textbf{0.574} & \textbf{69.4} & \textbf{70.5} & \textbf{0.578} & \textbf{68.5} \\\hline
\multicolumn{10}{|c|}{Multiple granularities   (IID+non-IID)}                                                                                                                                                        \\\hline
B-IUPC                                               & 53.8                     & 1.151          & -             & 70.5          & 0.589          & -             & 71.2          & 0.592          & -             \\\hline
B-MAA                                                & \underline{57.3}               & \underline{1.042}    & -             & 70.3          & 0.588          & -             & 71.4          & 0.573          & -             \\\hline
B-GS                                                 & 57.2                     & \underline{1.042}    & \underline{54.5}    & 70.2          & 0.593          & 68.3          & 71.1          & 0.585          & 68.5          \\\hline
R-GNNLM                                              & 54.4                     & 1.102          & -             & \underline{72.2}    & \underline{0.573}    & -             & \underline{72.1}    & \underline{0.568}    & -             \\\hline
B-M2A‡                                               & 58.7                     & 1.021          & 54.8          & 72.0           & 0.569          & 69.7          & 72.4          & 0.560          & 69.1          \\\hline
G-M2A‡                                               & 58.1                     & 1.074          & 53.6          & 70.4          & 0.598          & 67.4          & 71.8          & 0.569          & 68.4          \\\hline
R-M2A‡                                               & 60.3            & \textbf{0.954} & \textbf{56.7} & \textbf{73.7} & 0.548 & \textbf{70.7} & \textbf{74.2} & \textbf{0.535} & \textbf{71.4} \\ \hline
R-M2A                                                & \textbf{60.6} & 0.960          & 56.6          & 73.4          & \textbf{0.543}          & 70.2          & 73.6          & 0.545          & 70.9    \\ \hline      
\end{tabular}
\caption{Comparative test results for three personalized sentiment classification datasets (user and item attributes). The underline and backbone scores respectively meant the best scores in baselines and all models in each group. Our M2A models outperformed the previous works significantly ($p$\textless{}0.05) in Acc.}
\label{tab:2}
\end{table*}

\section{Experiments} \label{sec:4}
\subsection{Datasets and Evaluations} \label{sec:4.1}
We evaluate the proposed M2A frameworks on two types of prevalent datasets. (1) \textbf{Multi-domain sentiment analysis} includes \textbf{FDU-MTL}, which contains product and movie reviews across 16 domains from the category-attribute perspective~\cite{Liu2017}. (2) \textbf{Personalized sentiment analysis} includes \textbf{IMDB}, \textbf{Yelp-2013}, and \textbf{Yelp-2014}. These datasets exhibit data heterogeneities from user and item-attribute perspectives~\cite{Tang2015a}. Regarding annotated labels, FDU-MTL, IMDB, and Yelps are treated as a 2-class, 10-class, and 5-class classification problems, respectively.

For evaluation metrics, we adopt \textbf{Accuracy (Acc)} to measure the effectiveness of models for multi-domain sentiment analysis. For personalized sentiment analysis, we used \textbf{Acc} (primarily), \textbf{Rooted-Mean Square Error (RMSE)}, and \textbf{Macro-F1 (F1)} scores~\cite{Yuan2022,Zhang2023}.

\subsection{Experimental Settings} \label{sec:4.2}
We compared our M2A framework with previous state-of-the-art models across different datasets.

(1) For multi-domain datasets, the baselines include \textbf{BERT}~\cite{Devlin2019}, \textbf{DAEA}~\cite{Cai2019}, \textbf{BERTMaker}~\cite{Yuan2022}, \textbf{KCL-KB}~\cite{Yuan2023}, and \textbf{MTL}~\cite{Thrun1995}.

(2) For personalized datasets, the baselines include \textbf{variants of PLMs} in a coarse-grained view, as well as \textbf{IUPC}~\cite{Lyu2020}, \textbf{MAA}~\cite{Zhang2021}, \textbf{GS}~\cite{Zhang2023}, and \textbf{GNNLM}~\cite{Kertkeidkachorn2023} from multi-grained perspectives.

To investigate the effect of our M2A, we introduced diverse PLMs as backbones, including BERT (B)~\cite{Devlin2019}, RoBERTa (R)~\cite{Liu2019}, and GPT2 (G)~\cite{radfordlanguage}. We adopted FFT and PEFT for optimization, denoted as \textbf{M2A‡} and \textbf{M2A}, respectively. Here, M2A‡ provided a fair comparison with previous works also using FFT. \textbf{M2A†} represented the coarse-grained version jointed learned with M2A, handling non-IID-free scenarios, as discussed in §M2A Framework. All results were averaged over five runs. Detailed model configurations of hyperparameters can be found in the technical appendices.

\subsection{Comparative Results and Analysis}\label{sec:4.3}
Tables \ref{tab:1} and \ref{tab:2} reported the comparative results of the proposed M2A framework against previous works on multi-domain and personalized sentiment analysis datasets.

From Table~\ref{tab:1}, when individually trained for each domain, BERT achieved relatively the worst Acc scores. This phenomenon suggested that although PLMs could gain high performance in sentiment analysis, they continued encountering challenges due to data scarcity within each domain. When all domain datasets were gathered, multi-domain models achieved higher performance than BERT, indicating that more extensive and aggregated data could facilitate data-hungry neural networks to achieve generalized performance. B-MTL, which utilized multitask learning to handle all domains, achieved competitive results. However, it treated all samples as IID, disregarding data heterogeneities, which led to inferior performance compared to our B-M2A. The proposed M2A, from a Bayesian perspective, outperformed other baselines by ensemble multiple views and encoding robust domain features. Moreover, the results showed that PLMs using the same transformer structure exhibited varying performance with M2A, emphasizing the importance of selecting appropriate prior distributions ($p(w)$ and $p({\bf{x}}|w)$). The proposed M2A framework, which integrates both IID and non-IID features, consistently outperformed other models by effectively addressing data heterogeneities and leveraging domain-specific information.

\begin{table*}[!t]
\centering
\small
\begin{tabular}{|cl|c|c|c|c|}
\hline
\multicolumn{2}{|c|}{\textbf{Models}}                                                                                                                 & FDU-MTL & IMDB  & Yelp-2013 & Yelp-2014 \\ \hline
\multicolumn{2}{|c|}{B-M2A‡}                                                                                                                           & 93.86   & 58.70 & 72.00     & 72.44     \\ \hline
\multicolumn{1}{|c|}{\multirow{4}{*}{M2 Mixture}}      & - coarse view                                                                                 & 90.98   & -     & -         & -         \\ \cline{2-6} 
\multicolumn{1}{|c|}{}                                 & - fine-grained view (user)                                                                    & -       & 52.80 & 68.57     & 68.85     \\ \cline{2-6} 
\multicolumn{1}{|c|}{}                                 & - fine-grained view (item)                                                                    & -       & 58.43 & 71.42     & 71.58     \\ \cline{2-6} 
\multicolumn{1}{|c|}{}                                 & - fine-grained view (all)                                                                     & 92.78   & 52.21 & 67.65     & 67.65     \\ \hline
\multicolumn{1}{|c|}{\multirow{2}{*}{Module Learning}} & - text generation task                                                                        & 92.70   & 57.42 & 69.36     & 70.29     \\ \cline{2-6} 
\multicolumn{1}{|c|}{}                                 & \begin{tabular}[c]{@{}l@{}}- text generation task\\(except randomly masking)\end{tabular} & 93.56   & 57.41 & 69.80     & 71.02     \\ \hline
\multicolumn{2}{|c|}{G-M2A‡}                                                                                                                           & 92.92   & 58.10  & 70.41     & 71.79     \\ \hline
\multicolumn{1}{|c|}{\multirow{4}{*}{M2 Mixture}}      & - coarse view                                                                                 & 91.84   & -     & -         & -         \\ \cline{2-6} 
\multicolumn{1}{|c|}{}                                 & - fine-grained view (user)                                                                    & -       & 52.35 & 68.04     & 68.41     \\ \cline{2-6} 
\multicolumn{1}{|c|}{}                                 & - fine-grained view (item)                                                                    & -       & 57.25 & 70.35     & 71.49     \\ \cline{2-6} 
\multicolumn{1}{|c|}{}                                 & - fine-grained view (all)                                                                     & 92.86   & 51.46 & 67.63     & 67.97     \\ \hline
\multicolumn{1}{|c|}{Module Learning}                  & - text generation task                                                                        & 92.87   & 56.43 & 70.18     & 70.99     \\ \hline
\end{tabular}
\caption{The ablative test Acc on multi-domain and personalized sentiment analysis in terms of the mixture of multiple attributes and granularities (guided by Bayesian Inference) and module learning strategies (guided by BNN), respectively.}
\label{tab:3}
\end{table*}

As shown in Table~\ref{tab:2}, models integrating multiple granularities (from user and item-attribute perspectives) outperformed those considering only coarse-grained views. This demonstrated the importance of leveraging data heterogeneities collected from diverse sources during the downstream task adaptation of PLMs. Compared with GNNLM, which introduced a graph neural work to encode robust attribute or domain representations for improved performance, our M2A provided a multitask learning method based on BNN for the same purpose, resulting in competitive performance across all three datasets. 

Furthermore, we compared a general version of M2A (R-M2A†) against RoBERTa. The superior performance validated that our Bayesian learning approach, combined with joint learning, endowed M2A with a generalized capability to handle scenarios with unclear data heterogeneities. Consequently, R-M2A‡ obtained the best results on all three datasets across all three metrics, suggesting the advantage of our Bayesian learning-based M2A framework, especially as the backbone PLMs became more powerful.

\subsection{Ablation Study} \label{sec:4.4}
The test results of the ablation study on both multi-domain and personalized sentiment analysis are reported in Table~\ref{tab:3}. We used BERT and GPT-2 as backbones for these investigations.

First, we gradually eliminated attribute representations from coarse to fine-grained views. As different granularities were removed, the performance of both B-M2A‡ and G-M2A‡ declined to different extents. This indicated integrating multi-grained information enhanced final performance by mitigating the uncertainty of models. In personalized scenarios, eliminating user non-IID information had a more significant impact than item non-IID information, consistent with previous findings~\cite{Wu2018}. This was because user preferences tend to create more distinct non-IID distributions among texts than item characteristics, revealing the diversity among different attributes.

Next, we removed text generation tasks in the module learning strategy to investigate the effect of the prior likelihood estimator $p({\bf{x}}|w)$. The results for both B-M2A‡ and G-M2A‡ decreased with the removal of $p({\bf{x}}|w)$, indicating its significance. To explore the improvements of B-M2A‡ derived from $p({\bf{x}}|w)$ or 15\% tokens masking, we removed $p({\bf{x}}|w)$ while prevising the MLM targets and found that B-M2A‡ without $p({\bf{x}}|w)$ was worse than the above settings. This phenomenon further demonstrated the effect of $p({\bf{x}}|w)$ through BNN in Eq. (\ref{eq:4}).

\subsection{The Effect of Bayesian Learning} \label{sec:4.5}
An extensive experimental analysis was conducted to investigate how Bayesian learning theoretically facilitated our framework for text understanding.

\begin{table}[!t]
\centering
\small
\begin{tabular}{|cc|c|c|c|}
\hline
\multicolumn{2}{|c|}{\textbf{Methods}}                                       & \multirow{2}{*}{IMDB} & \multirow{2}{*}{Yelp-2013} & \multirow{2}{*}{Yelp-2014} \\ \cline{1-2}
\multicolumn{1}{|c|}{CGV}                  & FGV                    &                       &                            &                            \\ \hline
\multicolumn{1}{|c|}{\multirow{2}{*}{FFT}} & LoRA                   & OOM                   & OOM                        & OOM                        \\ \cline{2-5} 
\multicolumn{1}{|c|}{}                     & KronA                  & 58.60                 & 72.68                      & 73.52                      \\ \hline
\multicolumn{1}{|c|}{LoRA}                 & \multirow{2}{*}{KronA} & 59.35                 & 71.18                      & 72.22                      \\ \cline{1-1} \cline{3-5} 
\multicolumn{1}{|c|}{KronA}                &                        & 57.63                 & 68.03                      & 67.38                      \\ \hline
\end{tabular}
\caption{Dev Acc of R-M2A with different PEFTs for \{coarse\}+\{fine-grained\} views (denoted by CGV and FGV). LoRA has a low rank of 64, and KronA estimates a low-rank matrix using rank-one parameters. OOM represents out-of-memory in our settings due to large external parameters.}
\label{tab:4}
\end{table}

\begin{figure}[t!]
\centering
\includegraphics[width=2.5in]{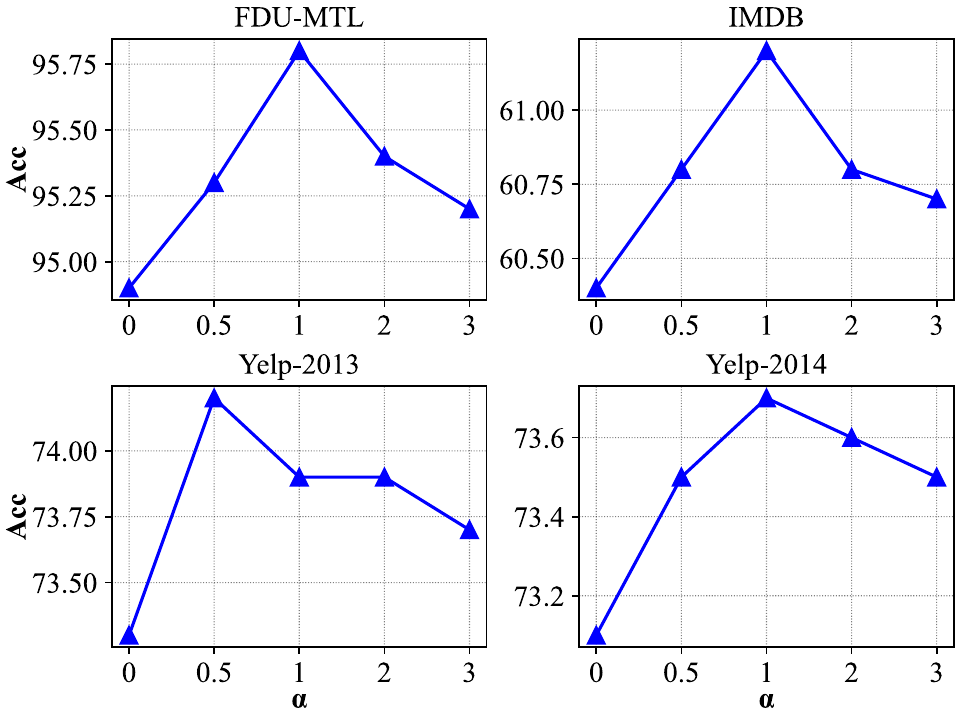}
\caption{Dev Acc of R-M2A with different multi-task.} \label{fig:3}
\end{figure}

\begin{table}[!t]
\centering
\small
\begin{tabular}{|l|c|}
\hline
\multicolumn{1}{|c|}{\textbf{B-M2A‡}} & FDU-MTL \\ \hline
w/o text generation                     & 92.70   \\ \hline
w/ labeled data                         & 93.86   \\ \hline
w/ unlabeled data                       & 93.65   \\ \hline
\end{tabular}
\caption{Test Acc of B-M2A‡ performing different text generalization tasks.}
\label{tab:5}
\end{table}

\noindent \textbf{Mixture of granularities.}
We initially employed different LoRA methods to evaluate the Bayesian learning-based mixture of multiple views. From Table~\ref{tab:4}, it was evident that the ascendancy of LoRA over KronA had a more pronounced effect on the final performance of R-M2A in coarse views than in fine-grained views. Consequently, LoRA was adopted to sample coarse models. Since KronA had much fewer parameters than LoRA and fine-grained views typically encompass many domains, we utilized KronA to instantiate domain modules in fine-grained views.

\noindent \textbf{Text generation tasks.}
We varied the balance factor $\alpha $ to explore the impact of text generation tasks in BNN assumptions. As shown in Figure~\ref{fig:3}, with the introduction of text generation tasks for optimizations, the final dev performance was improved, demonstrating the effect of text generation through Bayesian learning. The performance varied with different values of $\alpha $; models with either smaller or larger values $\alpha $ exhibited relatively lower Acc compared to those with an appropriate selection. This finding suggests caution when configuring the application in practice.

To further investigate the impact of text generation tasks, we introduced unlabeled data in each fine-grained domain to facilitate models to learn domain modules. As shown in Table~\ref{tab:5}, B-M2A‡ w/o text generation tasks represented models optimized with $p(y|{\bf{x}},w)$ on labeled data, and B-M2A‡ w/ unlabeled data represented models optimized with $p(y|{\bf{x}},w)$ on labeled data and $p({\bf{x}}|w)$ on unlabeled data. B-M2A‡ w/ unlabeled data model achieved better results than B-M2A‡ w/o generation tasks, indicating that unlabeled domain-specific data could also leverage domain-specific (non-IID) knowledge for fine-grained adaptations.

\noindent \textbf{Connection of granularities.}
We compared our joint learning strategy with other ensemble methods for integrating general view representation. From Table~\ref{tab:6}, R-M2A† achieved the best performance among all methods, demonstrating the effect of our proposed joint learning strategy guided by the Bayesian inference assumption. With the removal of KL or coarse-grained losses ${{\cal L}}_{{\rm{mtl}}}^{{\rm{N}}{{\rm{N}}^\dag }}$, the performance of R-M2A† degraded, revealing these losses could facilitate our framework to connect granularities. Moreover, R-M2A† outperformed R-M2A† w/o Sep, validating that ${w^{(ac')}}$ was required to continue being updated for a better generalization performance for IID-only scenarios.

\begin{table}[!t]
\centering
\tiny
\small
\begin{tabular}{|l|c|c|c|}
\hline
\textbf{Connections}                                            & IMDB & Yelp-2013 & Yelp-2014 \\ \hline
RoBERTa                                                         & 53.0 & 69.2      & 69.0      \\ \hline
R-M2A w/ Avg                                                    & 52.2 & 70.3      & 69.9      \\ \hline
R-M2A w/ Rand                                                   & 47.3 & 66.7      & 66.2      \\ \hline
R-M2A†                                                          & \textbf{54.2} & \textbf{70.7}      & \textbf{70.5}      \\ \hline
R-M2A† w/o Sep                                                  & 52.8 & 69.9      & 70.0      \\ \hline
R-M2A† w/o KL                                                   & 53.8 & 70.5      & 70.4      \\ \hline
R-M2A† w/o ${{\cal L}}_{{\rm{mtl}}}^{{\rm{N}}{{\rm{N}}^\dag }}$ & 53.8 & 70.3      & 70.1      \\ \hline
\end{tabular}
\caption{Test Acc of R-M2A with strategies of connecting granularities for IID-only scenarios. Avg means averaging all fine-grained modules for a general one. Rand randomly selected a fine-grained module for input data. Sep represents separation operation.}
\label{tab:6}
\end{table}

\section{Conclusions}
This study proposes an M2A framework to extract data heterogeneities from multi-source data for fine-grained adaptation. Our approach introduced a Bayesian analysis to rethink previous multi-domain-learning and attribute-injecting methods and provided a PEFT and joint learning strategy for facilitating PLMs’ adaptations. Experimental findings from multi-domain and personalized sentiment analysis showed that the proposed method could integrate models sampled from multiple attributes and granularities to eliminate data uncertainty. Moreover, the proposed method utilized BNN paradigms to leverage domain modules for performance improvements.

Future works intend to collect a multi-view dataset that contains more kinds of sources for further analysis and build an automatic data heterogeneity detector to verify the effectiveness of our methods.


\section*{Acknowledgments}
This work was supported by the National Natural Science Foundation of China (NSFC) under Grant Nos. 61966038, 62266051, and 62162068,
 the National Science and Technology Council (NSTC), Taiwan, ROC, under Grant No. 113-2221-E-155-046-MY3,
 and the Caiyun Postdoctoral Science Foundation under Grant No. C615300504090.
 The authors would like to thank the anonymous reviewers for their constructive comments.


\bibliography{custom}

\appendix
\newpage

\newpage

\section{Related Work}
\label{sec:A}
\subsection{Text Understanding} \label{sec:A.1}
Text understanding is fundamental and essential in natural language processing (NLP), such as sentiment analysis, topic labeling, question answering, and text classification~\cite{Zhang2018,Mao2012,Allam2012,kowsari2019text}. In the era of information exploration, it is desired to use machine learning algorithms to automate reliable text understanding for efficient information retrieval. Recently, PLMs initially pre-trained on a large corpus have provided robust prior knowledge and presented high-performance adaptations in downstream tasks~\cite{Qiu2020}. 

However, these PLMs are computation-hungry, especially with PLMs increasingly scaling large~\cite{,Liu2023}. FEFT provides an efficient and effective solution that fine-tunes several parameters and maintains the inherent generalization performance in PLMs. PEFT methods mainly contain two directions: 1) fine-tuning only a few parts of backbone parameters, such as BitFit~\cite{BenZaken2022}; 2) introducing lightweight parameters fine-tuned for downstream task adaptations, such as LoRA~\cite{Hu2021} and Prompt-Tuning~\cite{Lester2021b}. 

Inspired by~\citet{Edalati2022} and \citet{Hu2021}, we propose a KronA module to present a fine-grained view and a LoRA to present a coarse view. Finally, a simple sum strategy combines multiple views for PLM adaptations.

\subsection{Data Heterogeneity in Text} \label{sec:A.2}
Neural networks are data-hungry, leading to as much data as possible being collected from diverse sources, e.g., individual users, for robust generalization performance~\cite{Chen2021}. However, previous works seldom consider the heterogeneities in the crowdsourced data for fine-grained adaptations~\cite{Greff2017,Kim2014,Zulqarnain2020,Devlin2019}. Recently, multi-domain learning~\cite{Atzeni2020,Liu2018,Yang2015} and multi-attribute-injecting methods~\cite{Zhang2021,Amplayo2019,Wu2018} have provided a paradigm that learns a universal and domain-invariant representation combined with domain-specific features for fine-grained performance.

Although high performances have been gained, these methods do not provide a theoretical model design and optimization approach. In comparison, we improve current methods from a Bayesian perspective and introduce a unified framework leveraging pragmatic features for robust non-IID representation.

\subsection{Bayesian Learning} \label{sec:A.3}
Bayesian learning has gained considerable interest in addressing uncertainty via posterior uncertainty, generalizing while reducing overfitting~\cite{Hastings1970}, and sequential learning while retaining prior and past knowledge~\cite{Freitas2000}. The Bayesian learning methods closest to ours are Bayesian inference and Bayesian neural networks (BNNs), which inspire us for text understanding~\cite{Magris2023}. Based on the Bayesian diagram~\cite{Etz2018}, \citet{Chen2021} proposes a Bayesian model ensemble method to mitigate model drifts where each model is estimated from an individual client with individual annotated data. With the emergence of deep neural networks, the BNN requires the introduction of the prior distributions over model parameters such as parametric forms. This is because the prior distribution would influence the posterior distribution under Bayesian assumptions~\cite{Jospin2022}.

In contrast to previous works mainly based on the Monte Carlo method~\cite{Robert1999} to sample prior distributions, we utilized data heterogeneities in different views for uncertainty estimation.

\section{Connecting Previous Works} \label{sec:B}
\subsection{Dataset Distribution Perspective} \label{sec:B.1}
Non-IID information gathered from different sources might be across multiple attributes (denoted by ${{\cal A}}$) in practice. For each attribute $a \in {{\cal A}}$, the dataset ${{\cal D}}$ can be viewed from a coarse-grained and a fine-grained perspective, respectively.

\noindent \textbf{Coarse-grained view}.
Given a ${\rm{N}}{{\rm{N}}_{{w^{(ac)}}}}( \cdot )$ to estimate the inputs and the outputs, the coarse view means it equally treats all samples with IID. Based on Eq. (\ref{eq:3}), the predictive probabilities of current neural works for text understanding in a coarse-grained view can be defined as:
\begin{equation}
p(y|{\bf{x}};{{\cal D}}) \approx p(y|{\bf{x}};{w^{(ac)}}) \label{eq:11}
\end{equation}
where $S = \{ ac\} $.

\noindent \textbf{Fine-grained view.}
In a fine-grained view, given $|a{|_f}$ domains with models $\bigcup\nolimits_{as \in a} {{\rm{N}}{{\rm{N}}_{{w^{(as)}}}}} $ to estimate different domain datasets ${{{\cal D}}^{as}}$, the predictive probabilities in a fine-grained view can be defined as:
\begin{equation}
p(y|{\bf{x}};{{\cal D}}) \approx {1 \over {|a{|_f}}}\sum\limits_{as \in a} {p(y|{\bf{x}};{w^{(as)}})} \label{eq:12}
\end{equation}
where $S = a$ and ${w^{(as)}} \sim p({w^{(as)}}|{{\cal D}})$:
\begin{equation}
\begin{aligned}
p({w^{(as)}}|{{\cal D}}) &= p({w^{(as)}}|{{{\cal D}}^{(as)}}) \cdot p({{{\cal D}}^{(as)}}|{{\cal D}}) \cr
&\propto p({w^{(as)}}|{{{\cal D}}^{(as)}}) \label{eq:13}
\end{aligned}
\end{equation}

\begin{table*}[!t]
\centering
\small
\begin{tabular}{|l|c|c|c|c|c|c|}
\hline
\multicolumn{1}{|c|}{\textbf{Datasets}} & \#Samples & \#Domains in User & \#Domains in Item & \#Samples/user & \#Samples/item & \#Ratings \\\hline
IMDB                                  & 84919     & 1310   & 1635   & 64.82  & 51.94  & 10             \\\hline
Yelp-2013                             & 78966     & 1631   & 1633   & 48.42  & 48.36  & 5              \\\hline
Yelp-2014                             & 231163    & 4818   & 4194   & 47.97  & 55.11  & 5              \\\hline
\end{tabular}
\caption{The statistics of IMDB, Yelp-2013, and Yelp-2014 datasets from the user and item perspectives.}
\label{tab:7}
\end{table*}

\begin{table}[!t]
\centering
\small
\begin{tabular}{|l|c|c|c|c|}
\hline
\multicolumn{1}{|c|}{\textbf{Domains}} & Train & Dev & Test & Unlabeled \\\hline
Books                                & 1400  & 200 & 400  & 2000      \\\hline
Electronics                          & 1398  & 200 & 400  & 2000      \\\hline
DVD                                  & 1400  & 200 & 400  & 2000      \\\hline
Kitchen                              & 1400  & 200 & 400  & 2000      \\\hline
Apparel                              & 1400  & 200 & 400  & 2000      \\\hline
Camera                               & 1397  & 200 & 400  & 2000      \\\hline
Health                               & 1400  & 200 & 400  & 2000      \\\hline
Music                                & 1400  & 200 & 400  & 2000      \\\hline
Toys                                 & 1400  & 200 & 400  & 2000      \\\hline
Video                                & 1400  & 200 & 400  & 2000      \\\hline
Baby                                 & 1300  & 200 & 400  & 2000      \\\hline
Magazines                            & 1370  & 200 & 400  & 2000      \\\hline
Software                             & 1315  & 200 & 400  & 475       \\\hline
Sports                               & 1400  & 200 & 400  & 2000      \\\hline
IMDB                                 & 1400  & 200 & 400  & 2000      \\\hline
MR                                   & 1400  & 200 & 400  & 2000      \\\hline
\end{tabular}
\caption{The statistics of FDU-MTL dataset with 16 domains from the category perspective.}
\label{tab:8}
\end{table}

\subsection{Data Sample Perspective} \label{sec:B.2}
\textbf{Integration of multi-grained views.}
Each data sample $({\bf{x}},y)$ in the $a$-specific attribute can also be viewed from a coarse-grained and a fine-grained perspective, i.e., $({\bf{x}},y) \sim \bigcap\nolimits_{g \in {{\cal G}}} {{{{\cal D}}^{ag}}} $. As discussed in the previous sections (see §M2A Framework), $g \in {{\cal G}} = \{ c,f\} $, $af = {{\cal P}}{(({\bf{x}},y),a)_f}$, and $({\bf{x}},y)$ only belongs to ${af}$ fine-grained domain for simple settings, which does not hinder our implementations from being extended to fine-grained domain-overlapped scenarios. With the Bayesian inference assumption in Eq. (\ref{eq:3}), the predictive probabilistic of $({\bf{x}},y)$ can be formulated as:
\begin{equation}
\begin{aligned}
p(y|{\bf{x}};{{\cal D}}) &\approx {1 \over {\left| {{\cal G}} \right|}}\sum\limits_{g \in {{\cal G}}} {p(y|{\bf{x}};{w^{(ag)}})} \cr
&= p(y|{\bf{x}};\Omega (\bigcup\nolimits_{g \in {{\cal G}}} {{w^{(ag)}}} )) \label{eq:14}
\end{aligned}
\end{equation}
where ${w^{(ac)}} \sim p({w^{(ac)}}|{{\cal D}})$ and ${w^{(af)}} \sim p({w^{(af)}}|{{{\cal D}}^{(af)}})$. This inference is consistent with previous works utilizing multi-domain learning or attribute injection to incorporate IID and non-IID information via module integration, i.e., $\Omega ({w^{(ac)}},{w^{(af)}})$. These works optimize such unified models as:
\begin{equation}
\arg \max \prod\limits_{n = 1}^N {{\rm{Loss}}({y_n},{\rm{N}}{{\rm{N}}_{\Omega ({w^{(ac)}},{w^{(a{f_n})}})}}({{\bf{x}}_n}))} \label{eq:15}
\end{equation}
where $a{f_n} = {{\cal P}}{(({{\bf{x}}_n},{y_n}),a)_f}$; ${\rm{N}}{{\rm{N}}_{\Omega ({w^{(ac)}},{w^{(a{f_n})}})}}({{\bf{x}}_n})$ estimates $p(\Omega ({w^{(ac)}},{w^{(a{f_n})}})|({{{\cal D}}_x},{{{\cal D}}_y}))$ through $p({{{\cal D}}_y}|{{{\cal D}}_x},\Omega ({w^{(ac)}},{w^{(a{f_n})}}))$. Compared with previous works, the proposed method introduces an additional term $p({{{\cal D}}_x}|\Omega ({w^{(ac)}},{w^{(a{f_n})}}))$ for robust $p(\Omega ({w^{(ac)}},{w^{(a{f_n})}})|({{{\cal D}}_x},{{{\cal D}}_y}))$ estimation from a BNN perspective in Eq. (\ref{eq:9}).

\noindent \textbf{Integration of multi-attribute views.}
To present and ensemble attribute-specific representation on each data sample $({\bf{x}},y)$, Bayesian inference guides a module integration, including:
\begin{itemize}
\item Multi-attribute coarse-grained version
\begin{equation}
\begin{aligned}
p(y|{\bf{x}};{{\cal D}}) &\approx {1 \over {\left| {{\cal A}} \right|}}\sum\limits_{a \in {{\cal A}}} {p(y|{\bf{x}};{w^{(ac)}})} \cr
 &= p(y|{\bf{x}};\Omega (\bigcup\nolimits_{a \in {{\cal A}}} {{w^{(ac)}}} ))
\end{aligned} \label{eq:16}
\end{equation}

\item Multi-attribute multi-grained version
\begin{equation}
\begin{aligned}
p(y|{\bf{x}};{{\cal D}}) &\approx {1 \over {\left| {{\cal A}} \right|}}\sum\limits_{a \in {{\cal A}}} {({1 \over {\left| {{\cal G}} \right|}}\sum\limits_{g \in {{\cal G}}} {p(y|{\bf{x}};{w^{(ag)}})} )} \cr
&= {1 \over {\left| {{\cal A}} \right|\left| {{\cal G}} \right|}}\sum\limits_{a \in {{\cal A}},g \in {{\cal G}}} {p(y|{\bf{x}};{w^{(ag)}})} \cr
&= p(y|{\bf{x}};\Omega (\bigcup\nolimits_{a \in {{\cal A}},g \in {{\cal G}}} {{w^{(ag)}}} ))
\end{aligned} \label{eq:17}
\end{equation}
\end{itemize}

\begin{table*}[!]
\centering
\begin{tabular}{|l|c|c|}
\hline
\multicolumn{1}{|c|}{Models} & Predictions & Attention Maps    \\ \hline
R-M2A                      & 2 (\Checkmark)       & \multirow{3}{*}{\includegraphics[width=2.0in]{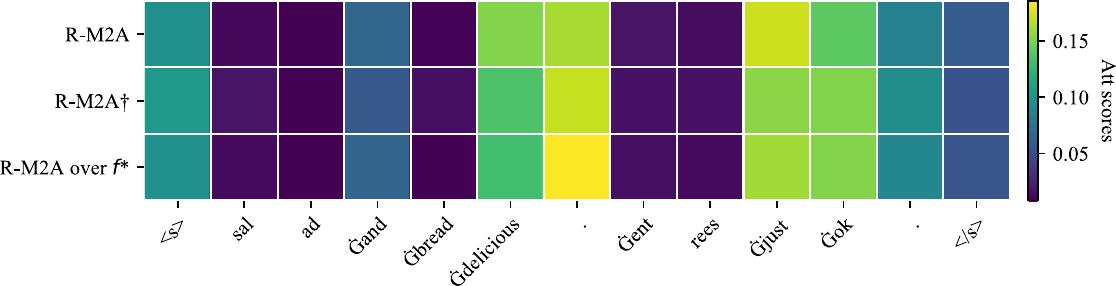}} \\ \cline{1-2}
R-M2A†                     & 3 (\XSolidBrush)     &                   \\ \cline{1-2}
R-M2A over $f^*$           & 3 (\XSolidBrush)     &                   \\ \hline
\end{tabular}
\caption{Attention visualization of the sample "salad and bread delicious. entrees just ok." with the golden rating of 2.}
\label{tab:9}
\end{table*}

\subsection{Connections among multiple granularities} \label{sec:B.3}
\textbf{From fine to coarse granularities.} 
Based on the Bayesian inference, the purpose of Eq.s (\ref{eq:11}) and (\ref{eq:12}) can be considered equivalent when drifts among sample models are mitigated, formally:
\begin{equation}
\begin{aligned}
p(y|{\bf{x}};{w^{(ac)}}) &= {1 \over {|a{|_f}}}\sum\limits_{as \in a} {p(y|{\bf{x}};{w^{(as)}})} \cr
&= p(y|{\bf{x}};\Omega (\bigcup\nolimits_{as \in a} {{w^{(as)}}} ))
\end{aligned}\label{eq:18}
\end{equation}

\noindent \textbf{Connection among fine and coarse granularities.}
To connect coarse and fine granularities, an external coarse-grained domain module ${w^{(ac)}}$ is introduced in Eq. (\ref{eq:18}), formally:
\begin{equation}
\small
\begin{aligned}
p(y&|{\bf{x}};{w^{(ac)}}) = {1 \over {|a{|_f}}}\sum\limits_{as \in a} {p(y|{\bf{x}};{w^{(as)}})} \cr
p(y&|{\bf{x}};{w^{(ac)}}) + p(y|{\bf{x}};{w^{(ac')}}) = \backslash \cr
&[{1 \over {|a{|_f}}}\sum\limits_{as \in a} {p(y|{\bf{x}};{w^{(as)}})} ] + p(y|{\bf{x}};{w^{(ac')}}) \cr
{1 \over 2}[p(y&|{\bf{x}};{w^{(ac)}}) + p(y|{\bf{x}};{w^{(ac')}})] = \backslash \cr
&{1 \over {|a{|_f}}}\sum\limits_{as \in a} {{1 \over 2}[p(y|{\bf{x}};{w^{(as)}}) + p(y|{\bf{x}};{w^{(ac')}})]} \cr
p(y&|{\bf{x}};\Omega ({w^{(ac)}},{w^{(ac')}})) = \backslash \cr
&{1 \over {|a{|_f}}}\sum\limits_{as \in a} {[p(y|{\bf{x}};\Omega ({w^{(as)}},{w^{(ac')}}))]} \cr
{1 \over {|a{|_f}}}&\sum\limits_{as \in a} {[p(y|{\bf{x}};\Omega ({w^{(ac)}},{w^{(ac')}}))]} = \backslash \cr
&{1 \over {|a{|_f}}}\sum\limits_{as \in a} {[p(y|{\bf{x}};\Omega ({w^{(as)}},{w^{(ac')}}))]}
\end{aligned} \label{eq:19}
\end{equation}
Regarding each data sample $({\bf{x}},y) \sim \bigcap\nolimits_{g \in {{\cal G}}} {{{{\cal D}}^{ag}}}$ and $g = \{ c,c',f\} $ it can get:
\begin{equation}
p(y|{\bf{x}};\Omega ({w^{(ac)}},{w^{(ac')}})) = p(y|{\bf{x}};\Omega ({w^{(af)}},{w^{(ac')}})) \label{eq:20}
\end{equation}
This results in the introduction of a joint learning strategy for improving generalization performance. Additionally, the combination of results in Eq.s (\ref{eq:18}) and (\ref{eq:19}) can also induce a general inference for coarse-grained scenario (i.e., R-M2A w/ Avg, as discussed in §Experiments), formally:
\begin{equation}
\begin{aligned}
p(y&|{\bf{x}};\Omega ({w^{(ac)}},{w^{(ac')}})) \cr
&= {1 \over {|a{|_f}}}\sum\limits_{as \in a} {[p(y|{\bf{x}};\Omega ({w^{(as)}},{w^{(ac')}}))]} \cr
&= p(y|{\bf{x}};\Omega (\Omega (\bigcup\nolimits_{as \in {{a}}} {{w^{(as    )}}} ),{w^{(ac')}}))
\end{aligned} \label{eq:21}
\end{equation}
R-M2A w/ Avg has been on par with RoBERTa directly fine-tuned from IID scenarios.

\section{Detailed Experimental Settings} \label{sec:C}
\textbf{Detailed datasets.} \label{sec:C.1}
Detailed multi-domain and personalized sentiment analysis datasets statistics are listed in Tables~\ref{tab:7} and \ref{tab:8}.

\noindent \textbf{Detailed hyperparameters.}
Regarding coarse-grained modules in transformer layers, they were applied for query, value, and intermediate terms with the low-rank factor $r$ of 128, which is grid searched over [32, 64, 128, 256]). Fine-grained modules were applied for only query and value terms with the low-rank factor ($r_{out}$, $r_{in}$) of (32, 24) selected over [(1, 768), (24, 32), (48, 16), (16,48) (32, 24), (768, 1)] according to~\cite{Edalati2022}. Following the previous works~\cite{Hu2021}, we initialized all domain modules (both coarse and fine-grained ones) with $A$ (or $C$) of uniform distribution and $B$ (or $D$) of zeroed distribution, which could preserve the primitive capability of PLMs. We set the loss coefficient of 0.5 for text generation and coarse-grained costs during optimization and default one of 1 for others. The optimizer was AdamW, with a learning rate 2e-5 and a gradient clip of 2. All experiments were conducted in one 3090 (24G) GPU device.

\section{Case Study} \label{sec:D}
\textbf{Visualizations across M2 features.}
To validate our observations and demonstrate the effectiveness of $p({{{\cal D}}_x}|w)$ and domain modules, we randomly take one review instance in the Yelp-2013 dataset for example. We visualize attention scores (multi-head averaged) over the token of \texttt{<s>} at the last transformer layer in R-M2A, R-M2A over $f^*$, and R-M2A†, shown in Table~\ref{tab:9}. R-M2A over $f^*$ means randomly selecting another user module to replace the exact one $f$ for the sampled instance. Note that the lighter color indicates higher attention weights. For the review from PLM to R-M2A, the attention scores have been updated to estimate the final predictions $p(y|{\bf{x}};{{\cal D}})$ with correct predictions. Without the $p({{{\cal D}}_x}|w)$ optimization purpose, attention scores of M2A† changed and led to a bias prediction. Moreover, when the exact user module is modified, M2A over $f^*$ produces a different attention score, demonstrating the effectiveness of user domain representation for helping PLMs locate non-IID information.

\end{document}